\documentclass{article}
\usepackage[letterpaper]{geometry}
\usepackage{hyperref}
\usepackage[utf8]{inputenc}
\usepackage[T1]{fontenc}
\usepackage{amsmath, amssymb, amsfonts, mathtools, stmaryrd}
\usepackage{mleftright}
\usepackage{varwidth}     
\usepackage{algorithm}
\usepackage[noend]{algpseudocode}
\usepackage[table]{xcolor}
\usepackage{tabularx}

\usepackage[most]{tcolorbox}
\usepackage{float}
\usepackage{arydshln}
\newtcolorbox{solidrowbox}[1][]{colback=white, colframe=green!60!black,
  boxrule=0.8pt, arc=3pt, boxsep=0pt, left=2pt, right=2pt, top=2pt, bottom=2pt,
  sharp corners=south, enhanced, #1}

\usepackage{etoolbox}
\usepackage{amsthm}
\theoremstyle{plain}

\theoremstyle{definition}

\theoremstyle{remark}

\usepackage{tikz}
\usepackage{pgfplots, pgfplotstable}
\usetikzlibrary{arrows, decorations.text, positioning, scopes, shapes, patterns}

\usepackage[table]{xcolor}
\usepackage[most]{tcolorbox}

\usepackage{url}
\usetikzlibrary{calc} 
\input{tangocolors.sty}  
\usetikzlibrary{arrows.meta}
\usepackage{subcaption}
\usepackage{tabularx}
\usepackage{natbib}
\usepackage{caption}
\DeclareCaptionLabelFormat{algnonumber}{Algorithm}
\DeclareCaptionSubType*{algorithm}

\DeclareCaptionLabelFormat{alglabel}{Alg.~#2}

\def\withcolors{1}
\def\withnotes{1}

\ifnum\withcolors=1

  \newcommand{\mcolor}[1]{{\color{orange} {#1}}}
  \newcommand{\vcolor}[1]{{\color{green} {#1}}}
  
\else

\fi

\ifnum\withnotes=1
  \newcommand{\mnote}[1]{\par\mcolor{\textbf{PM: }\sf #1}}
  \newcommand{\vnote}[1]{\par\vcolor{\textbf{VR: }\sf #1}}
  \newcommand{\cnote}[1]{\par\jcolor{\textbf{MC: }\sf #1}}
  \newcommand{\snote}[1]{\par\jcolor{\textbf{SN: }\sf #1}}
  
\else
  \newcommand{\mnote}[1]{}
  \newcommand{\vnote}[1]{}
  \newcommand{\cnote}[1]{}
  \newcommand{\snote}[1]{}
  
\fi
\newcommand{\ignore}[1]{\leavevmode\unskip}

\newcommand{\uRoman}[1]{\uppercase\expandafter{\romannumeral#1}}

\def\BibTeX{{\rm B\kern-.05em{\sc i\kern-.025em b}\kern-.08em
    T\kern-.1667em\lower.7ex\hbox{E}\kern-.125emX}}

\allowdisplaybreaks

\begin{document}

\title{Atlas Urban Index: A VLM-Based Approach for Spatially and Temporally Calibrated Urban Development Monitoring}

\author{ Mithul Chander \\ Propheus\\ \url{mithul@propheus.com} \and Sai Pragnya Ranga$^*$\\Propheus\\ \url{saipragnyaranga@gmail.com}  \and Prathamesh Mayekar\\ Propheus\\\url{prathamesh@propheus.com}  }
\date{}
\maketitle

\begin{abstract}
   We introduce the  {\em Atlas Urban Index} (AUI), a metric for measuring urban development computed using Sentinel-2 \citep{spoto2012sentinel2} satellite imagery. Existing approaches, such as the {\em Normalized Difference Built-up Index} (NDBI), often struggle to accurately capture urban development due to factors like atmospheric noise, seasonal variation, and cloud cover. These limitations hinder large-scale monitoring of human development and urbanization. To address these challenges, we propose an approach that leverages {\em Vision-Language Models }(VLMs) to provide a development score for regions. Specifically, we collect a time series of Sentinel-2 images for each region. Then, we further process the images within fixed time windows to get an image with minimal cloud cover, which serves as the representative image for that time window. To ensure consistent scoring, we adopt two strategies: (i) providing the VLM with a curated set of reference images representing different levels of urbanization, and (ii) supplying the most recent past image to both anchor temporal consistency and mitigate cloud-related noise in the current image. Together, these components enable AUI to overcome the challenges of traditional urbanization indices and produce more reliable and stable development scores. Our qualitative experiments on Bangalore suggest that AUI outperforms standard indices such as NDBI.
   
   
  
\end{abstract}
\section*{Keywords}
Atlas Urban Index (AUI), Vision-Language Models (VLMs), Sentinel-2 satellite imagery, Normalized Difference Built-up Index (NDBI).

\renewcommand{\thefootnote}{}%
\footnotetext{$^*$Work done while the author was affiliated with Propheus.}
\addtocounter{footnote}{0}

\renewcommand{\thefootnote}{}%
\footnotetext{An abridged version of this paper will be presented at and appear in the \textit{Proceedings of ACM IKDD CODS 2025.}}%
\addtocounter{footnote}{0}

\renewcommand{\thefootnote}{\arabic{footnote}}
\section{Introduction}

Global urbanization is accelerating at an unprecedented pace, with the share of people living in cities expected to rise from 
$\sim 55\%$ today to nearly $70\%$ by $2050$~\citep{unurban2025}. Much of this growth will be concentrated in rapidly developing countries such as India, China, and Nigeria. In this context, developing automated mechanisms for identifying areas that are urbanizing quickly has become increasingly important for both businesses and governments. 

Real estate developers depend on reliable measures of growth to guide site selection and monitor projects, while retailers and restaurants use urban expansion patterns both to identify promising markets and to monitor the performance of existing stores. Government agencies, meanwhile, require robust monitoring tools to evaluate urban policy outcomes, allocate resources effectively, and ensure sustainable city planning.

Across all these domains, consistent and accurate tracking of urban development forms the foundation for data-driven decision-making in rapidly evolving environments. This need becomes even more critical when development occurs through informal construction, where government records are absent or incomplete and independent monitoring is the only viable means of tracking change.

In this paper, we propose the \emph{Atlas Urban Index} (AUI), a mechanism for measuring urban development using satellite imagery and {\em Vision-Language Models} (VLMs). In our demonstration, we compute development scores across a standardized spatial unit, \emph{Geohash 5} (GH5)\footnote{GH5 is a standardized grid format that divides the Earth’s surface into cells of approximately $4.89$ km× $4.89$ km; see \citep{geohash2025} for a detailed description.}, using Sentinel-2 imagery~\cite{spoto2012sentinel2} in combination with OpenAI’s GPT-4o-mini~\cite{openai2024gpt4omini}. The proposed mechanism is general in the sense that it can be applied to other standardized spatial units, provided that high-resolution satellite imagery is available at the corresponding spatial scale. It can also be extended to other sources of spatial imagery and paired with alternative VLMs.

Our approach involves collecting a time series of Sentinel-2 images for each region at fixed intervals. From these, we select the images with the least cloud cover to ensure quality. In our demonstration, we focus on Bangalore, assembling images from the past ten years and selecting one image every six months. To generate consistent development scores, we adopt two strategies. First, we provide the VLM with a curated set of reference images that represent different levels of urbanization, which helps calibrate its scoring spatially. Second, we supply the image from the previous interval alongside the current one, which helps calibrate its scoring temporally and mitigate the effects of cloud-related noise. Together, these strategies allow the Atlas Urban Index (AUI) to overcome the limitations of traditional urbanization indices and deliver more reliable, stable measures of development.

We validate AUI on two Bangalore regions: the rapidly developing airport area and the more modestly growing Bannerghatta National Park region. In both cases, AUI aligns with known urbanization trends, whereas NDBI fails to reliably capture development.

\section{AUI Computation Pipeline}

We break down the computation of the AUI into two main steps. First, we describe the data processing stage, where satellite imagery is collected and prepared for analysis. Second, we detail the VLM-based scoring pipeline, which assigns development scores based on both reference images and temporal context.

\subsection{Data Processing}
For each region of interest, we obtain Sentinel-2~\cite{spoto2012sentinel2} imagery corresponding to the first and third quarters of each year. Within each quarter, the available images are filtered by cloud cover, ensuring that the least cloudy image is selected for analysis. This filtering is essential given the high sensitivity of remote sensing indices such as NDBI to atmospheric noise and cloud-related artifacts. To maintain an adequate temporal buffer, we omit the second and fourth quarters, resulting in one representative image every six months.

The selected image is stored in TIFF format~\cite{wiggins2001image}, which contains all $13$ spectral bands captured by Sentinel-2. To prepare inputs for the vision-language model (VLM), we extract the visible RGB bands ($B4, B3, B2$) and convert them into JPEG format. These RGB composites serve as the primary inputs to the scoring pipeline. Unless otherwise specified, all images discussed hereafter are generated using this procedure. Reference images are processed in the same way, but they are manually selected and assigned discrete AUI scores representing varying levels of urbanization.

\subsection{VLM-Based Scoring Pipeline}

\begin{figure}
    \centering
    
    \begin{subfigure}[b]{0.4\linewidth}
        \centering
        \includegraphics[width=\linewidth]{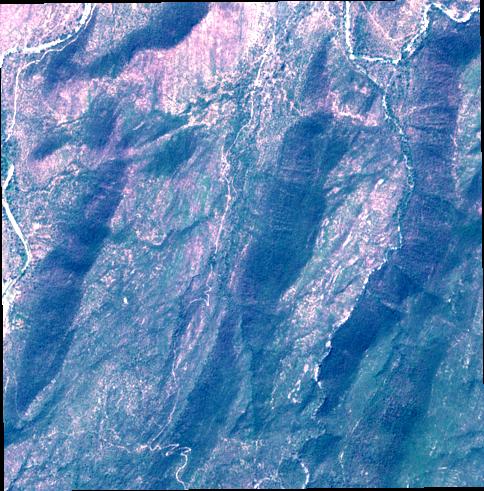}
        \caption{AUI Range 0}
    \end{subfigure}
    \hfill
    \begin{subfigure}[b]{0.4\linewidth}
        \centering
        \includegraphics[width=\linewidth]{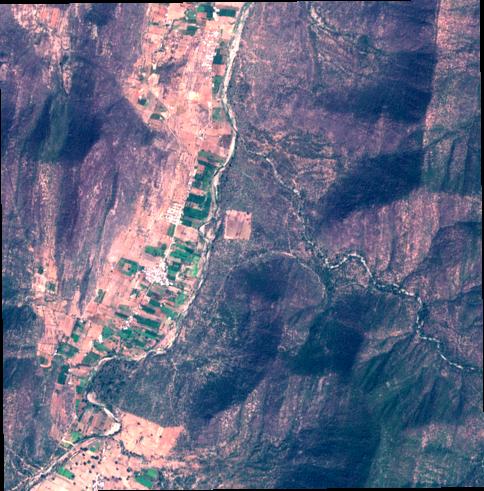}
        \caption{AUI Range 1--2}
    \end{subfigure}
    
    \vspace{0.5cm}
    
    \begin{subfigure}[b]{0.4\linewidth}
        \centering
        \includegraphics[width=\linewidth]{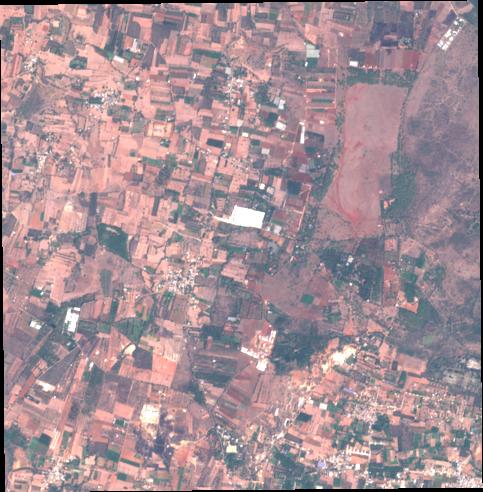}
        \caption{AUI Range 3--4}
    \end{subfigure}
    \hfill
    \begin{subfigure}[b]{0.4\linewidth}
        \centering
        \includegraphics[width=\linewidth]{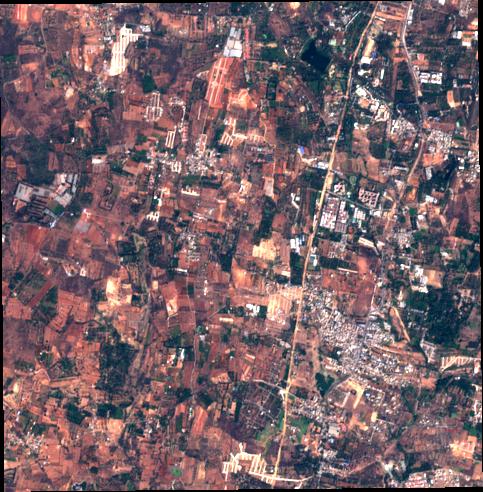}
        \caption{AUI Range 5--6}
    \end{subfigure}
    
    \vspace{0.5cm}
    
    \begin{subfigure}[b]{0.4\linewidth}
        \centering
        \includegraphics[width=\linewidth]{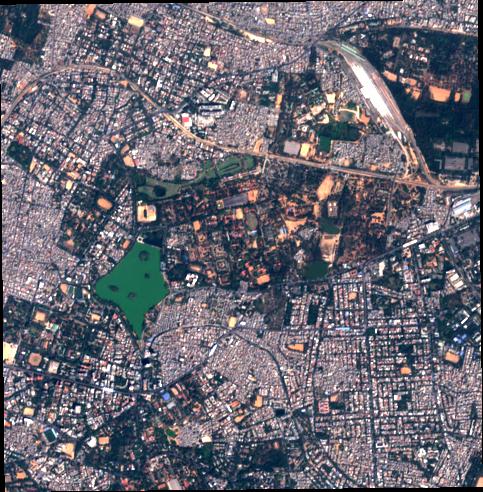}
        \caption{AUI Range 7--8}
    \end{subfigure}
    \hfill
    \begin{subfigure}[b]{0.4\linewidth}
        \centering
        \includegraphics[width=\linewidth]{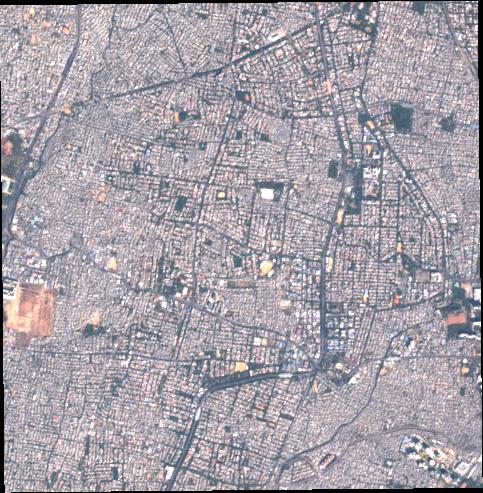}
        \caption{AUI Range 9--10}
    \end{subfigure}
    
    \caption{Reference images corresponding to AUI ranges.}
    \label{fig:1_aui_bins}
\end{figure}

Once the images are prepared, we perform scoring using a vision-language model (VLM), in our case, GPT-4o mini~\cite{openai2024gpt4omini}. A na\"{\i}ve approach to using the VLM would be to simply supply it with the image and prompt it to assign an urban development score. However, the challenge with this approach lies in the unpredictability of VLM behavior: small variations in the input image do not guarantee proportionally consistent changes in the output. As a result, two key issues arise -- {\em Spatial Miscalibration}, where a region with higher urban development may receive a lower score than one with less development, and {\em Temporal Miscalibration}, where the same region may be assigned a lower score at a later date despite visible urban growth, or conversely, a higher score despite a decline in urbanization. We address these calibration issues by providing the VLM with three inputs. \begin{enumerate}
    \item  The current image of the region for which the AUI is to be computed.
    \item A manually curated set of reference images representing varying levels of urbanization and their respective AUI\footnote{In future applications extending beyond Bangalore, it would be instructive to expose the VLM to regionally diverse examples that share similar development levels (e.g., a desert city in the Middle East and a tropical city in Southeast Asia), ensuring cross-context consistency. For the present demonstration, since all reference regions belong to the Bangalore metropolitan area and share similar environmental characteristics, such diversification is not required.}: Rajajinagar (GH5: \textit{tdr1u}, AUI: $9$–$10$), Halasuru (GH5: \textit{tdr1y}, AUI: $7$–$8$), Kaggalipura (GH5:  \textit{tdr0g}, AUI: $5$–$6$), Hesaraghatta (GH5: \textit{tdr4f}, AUI: $3$–$4$), Jallipalya (GH5: \textit{tdp54}, AUI: $1$–$2$), and a forest region near Koothampalyam on the Tamil Nadu–Karnataka border (GH5: \textit{tdp49}, AUI: $0$); see Figure~\ref{fig:1_aui_bins}.
    \item  The image of the same region from the immediate past time period along with its assigned AUI.  Using the past image allows us to mitigate fluctuations caused by transient factors such as seasonal variation or cloud-related artifacts.
\end{enumerate}

\section{Examples}
In our examples, we compare AUI with another widely used metric for urban development, the Normalized Difference Built-up Index (NDBI). For a Sentinel-2 image, the NDBI at the pixel level is given by
\[
\text{NDBI} = \frac{\text{SWIR} - \text{NIR}}{\text{SWIR} + \text{NIR}},
\]
where $\text{SWIR}$ denotes the shortwave infrared band (B11) and $\text{NIR}$ denotes the near-infrared band (B8). To convert pixel-level NDBI values to an image-level score, we simply take the average over all pixels.

\begin{figure}[h]
    \centering
    
    \begin{subfigure}[b]{0.4\linewidth}
        \centering
        \includegraphics[width=\linewidth]{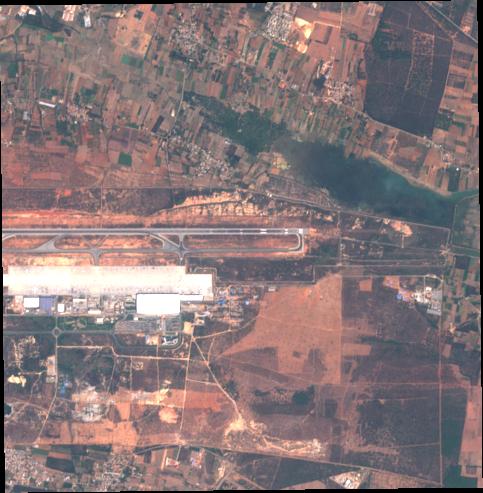}
        \caption{2016-01-01}
    \end{subfigure}
    \hfill
    \begin{subfigure}[b]{0.4\linewidth}
        \centering
        \includegraphics[width=\linewidth]{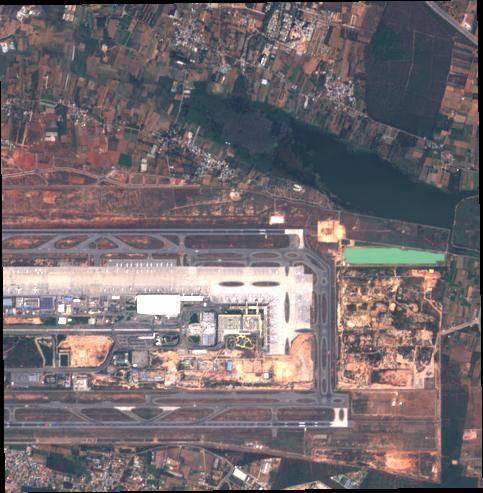}
        \caption{2025-01-01}
    \end{subfigure}
    
    \caption{Comparison of Kempegowda International Airport development between 2016 and 2025.}
    \label{fig:airport_comparission}
\end{figure}

\begin{figure}[H]
    \centering
    \begin{tikzpicture}
        \begin{axis}[
            width=0.8\linewidth,
            height=6cm,
            grid=major,
            xlabel={Date},
            ylabel={AUI},
            title={AUI},
                 xlabel style={yshift=-15pt}, 
            xtick={1,3,5,7,9,11,13,15,17,19}, 
            xticklabels={
                2016-01,2017-01,2018-01,2019-01,
                2020-01,2021-01,2022-01,2023-01,
                2024-01,2025-01
            },
            x tick label style={rotate=45, anchor=east},
            mark=*,
        ]
        \addplot[
            color=blue,
            thick,
            mark=*,
            ] coordinates {
                (1, 7.2)
                (2, 7.4)
                (3, 7.4)
                (4, 7.4)
                (5, 7.4)
                (6, 7.6)
                (7, 7.6)
                (8, 7.6)
                (9, 7.6)
                (10, 7.8)
                (11, 7.8)
                (12, 7.7)
                (13, 7.8)
                (14, 7.9)
                (15, 8.0)
                (16, 8.0)
                (17, 8.0)
                (18, 8.2)
                (19, 8.2)
            };
        \end{axis}
    \end{tikzpicture}
    \caption{AUI for GH5 \textit{tdr70} from 2016-2025}
    \label{fig:airport_aui}
\end{figure}
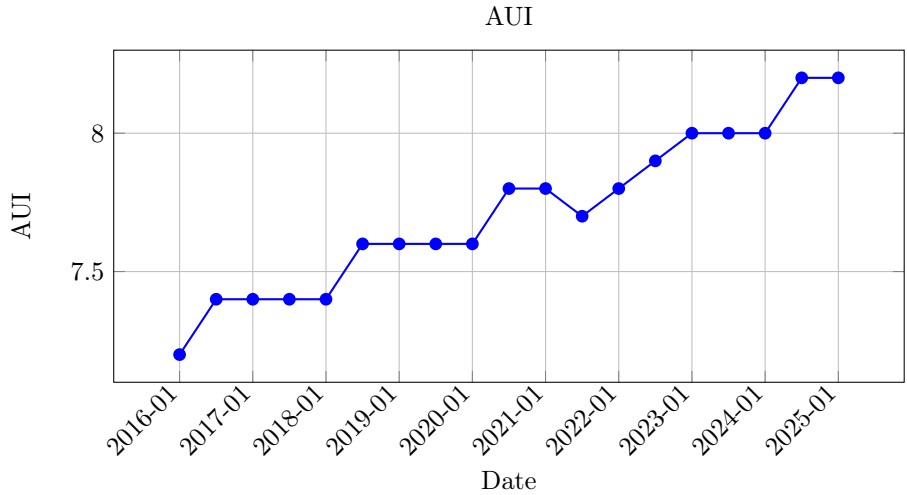

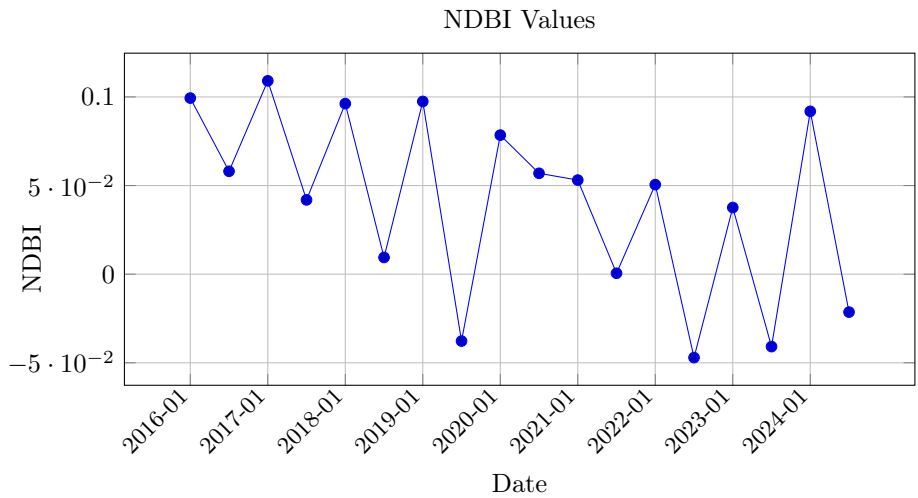
\begin{figure}[H]
    \centering
    \begin{tikzpicture}
        \begin{axis}[
            width=0.8\linewidth,
            height=6cm,
            grid=major,
            xlabel=Date,
            ylabel=NDBI,
            title={NDBI Values},
            xlabel style={yshift=-15pt}, 
             ylabel style={xshift=-15pt},
            xticklabel style={rotate=45, anchor=east, font=\small},
            symbolic x coords={
                2016-01,2016-07,
                2017-01,2017-07,
                2018-01,2018-07,
                2019-01,2019-07,
                2020-01,2020-07,
                2021-01,2021-07,
                2022-01,2022-07,
                2023-01,2023-07,
                2024-01,2024-07
            },
            xtick={
                2016-01,2017-01,2018-01,2019-01,
                2020-01,2021-01,2022-01,2023-01,
                2024-01
            }
        ]
            \addplot coordinates {
                (2016-01,0.099356)
                (2016-07,0.058056)
                (2017-01,0.109074)
                (2017-07,0.041918)
                (2018-01,0.096219)
                (2018-07,0.009447)
                (2019-01,0.097447)
                (2019-07,-0.037739)
                (2020-01,0.078452)
                (2020-07,0.056913)
                (2021-01,0.053083)
                (2021-07,0.000521)
                (2022-01,0.050523)
                (2022-07,-0.047060)
                (2023-01,0.037600)
                (2023-07,-0.040851)
                (2024-01,0.091858)
                (2024-07,-0.021398)
            };
        \end{axis}
    \end{tikzpicture}
    \caption{NDBI for GH5 \textit{tdr70} from 2016-2025}
    \label{fig:airport_ndbi}
\end{figure}

The temporal analysis of \textit{tdr70} using Sentinel-2 satellite imagery between $2016$ and $2025$ highlights the region’s progressive development. As shown in Figure \ref{fig:airport_aui}, the AUI  effectively capture the visible urban growth in the imagery, rising steadily from $7.2$ in early $2016$ to $8.2$ by $2025$. In contrast, Figure \ref{fig:airport_ndbi} demonstrates that  NDBI  fails to reflect this transformation.

Figure \ref{fig:airport_aui} details how the change in AUI over the ten-year period closely mirrors the physical development observed in the region; refer to Appendix  \ref{s:A} for all the satellite  images for region \textit{tdr70} over the ten-year period. In more detail:
\begin{itemize}
\item Between $2016$ and $2019$, only the basic foundation of the airport is visible, marking the initial phase of urbanization, with AUI increasing from $7.2$ to $7.4$.
\item Between $2019$ and $2022$, AUI rises to $7.8$, corresponding to the completion of Terminal~$2$, expansion of road networks, emergence of commercial infrastructure, and reduction in agricultural land.
\item Between $2022$ and $2025$, AUI climbs further to $8.2$, reflecting the establishment of dense commercial infrastructure and a mature urban form.
\end{itemize}

\subsection{Example 2}
Next, we analyze the development of Bannerghatta National Park in Bangalore, India, between $2016$ and $2025$. We focus on GH5 \textit{tdr0t} within this region. Initially, this area consisted largely of natural landscapes and exhibited only limited urban expansion over the ten-year period.

\begin{figure}[H]
    \centering
    \begin{tikzpicture}
        \begin{axis}[
            width=0.8\linewidth,
            height=6cm,
            grid=major,
            xlabel={Date},
            ylabel={AUI},
                 xlabel style={yshift=-15pt}, 
            xtick={1,3,5,7,9,11,13,15,17,19}, 
            xticklabels={
                2016-01,2017-01,2018-01,2019-01,
                2020-01,2021-01,2022-01,2023-01,
                2024-01,2025-01
            },
            x tick label style={rotate=45, anchor=east},
            mark=*,
        ]
        \addplot[
            color=blue,
            thick,
            mark=*,
            ] coordinates {
                (1, 2.5)
                (2, 2.5)
                (3, 2.6)
                (4, 2.6)
                (5, 2.7)
                (6, 3.0)
                (7, 3.5)
                (8, 3.7)
                (9, 3.7)
                (10, 3.7)
                (11, 3.9)
                (12, 3.9)
                (13, 3.9)
                (14, 3.9)
                (15, 3.9)
                (16, 3.9)
                (17, 4.1)
                (18, 4.1)
                (19, 4.1)
            };
        \end{axis}
    \end{tikzpicture}
    \caption{AUI for GH5 \textit{tdr0t} region from 2016--2025}
    \label{fig:bannerghatta_aui}
\end{figure}
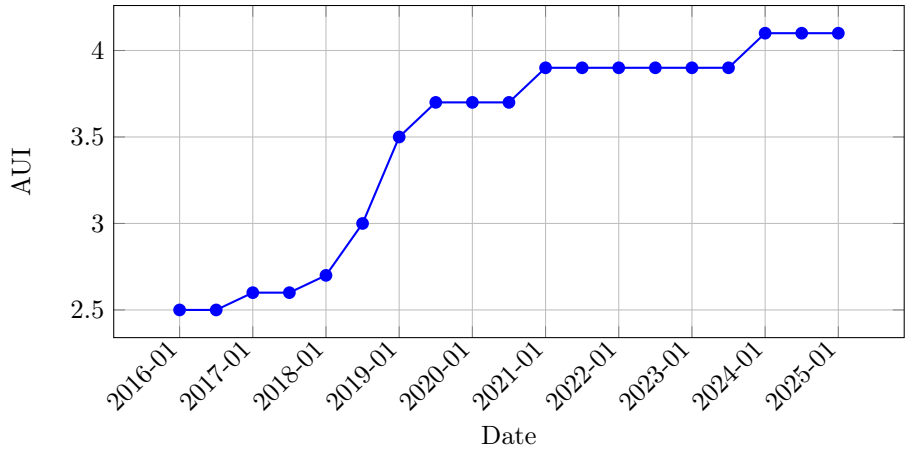

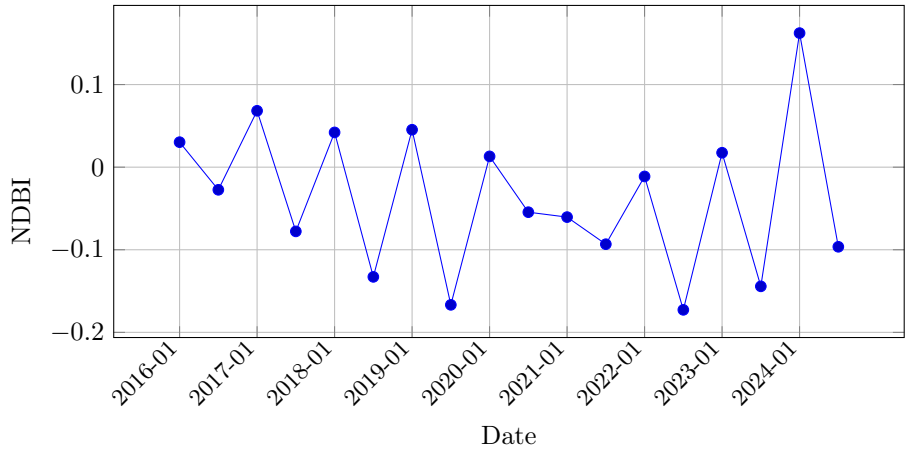
\begin{figure}[h]
    \centering
    \begin{tikzpicture}
        \begin{axis}[
            width=0.8\linewidth,
            height=6cm,
            grid=major,
            xlabel=Date,
            ylabel=NDBI,
               xlabel style={yshift=-15pt}, 
             ylabel style={xshift=-15pt},
            xticklabel style={rotate=45, anchor=east, font=\small},
            symbolic x coords={
                2016-01,2016-07,
                2017-01,2017-07,
                2018-01,2018-07,
                2019-01,2019-07,
                2020-01,2020-07,
                2021-01,2021-07,
                2022-01,2022-07,
                2023-01,2023-07,
                2024-01,2024-07
            },
            xtick={
                2016-01,2017-01,2018-01,2019-01,
                2020-01,2021-01,2022-01,2023-01,
                2024-01
            }
        ]
            \addplot coordinates {
                (2016-01,0.030267)
                (2016-07,-0.027413)
                (2017-01,0.068244)
                (2017-07,-0.077752)
                (2018-01,0.042066)
                (2018-07,-0.132821)
                (2019-01,0.045319)
                (2019-07,-0.166687)
                (2020-01,0.013057)
                (2020-07,-0.054505)
                (2021-01,-0.060465)
                (2021-07,-0.093263)
                (2022-01,-0.011229)
                (2022-07,-0.172748)
                (2023-01,0.017461)
                (2023-07,-0.144293)
                (2024-01,0.162315)
                (2024-07,-0.096404)
            };
        \end{axis}
    \end{tikzpicture}
    \caption{NDBI for GH5 \textit{tdr0t} from 2016-2025}
    \label{fig:bannerghatta_ndbi}
\end{figure}

The temporal analysis of \textit{tdr0t} using Sentinel-2 satellite imagery highlights the region’s modest development. As shown in Figure~\ref{fig:bannerghatta_aui}, the AUI captures this modest urban growth in the imagery, increasing from $2.5$ in early $2016$ to $4.1$ by $2025$. In contrast, and consistent with the previous example, Figure~\ref{fig:bannerghatta_ndbi} shows that the NDBI fails to reflect this transformation.


\begin{figure}[h]
    \centering
    
    \begin{subfigure}[b]{0.42\linewidth}
        \centering
        \includegraphics[width=\linewidth]{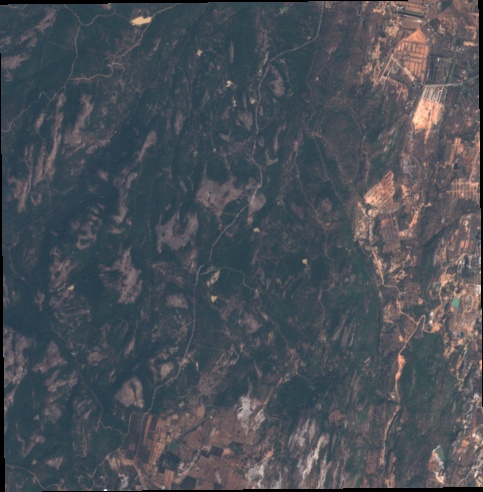}
        \caption{2016-01-01}
    \end{subfigure}
    \hfill
    \begin{subfigure}[b]{0.42\linewidth}
        \centering
        \includegraphics[width=\linewidth]{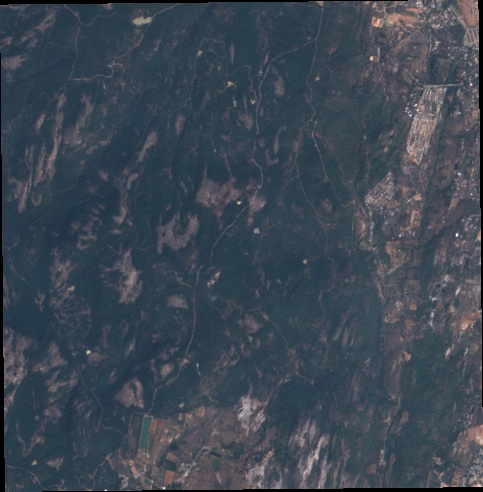}
        \caption{2025-01-01}
    \end{subfigure}
    
    \caption{Comparison of development around Bannerghatta National Park between 2016 and 2025.}
    \label{fig:bannerghatta_comparison}
\end{figure}

Figure \ref{fig:bannerghatta_aui} details how the change in AUI over the ten-year period closely mirrors the physical development observed in the region;  refer to Appendix \ref{s:B} for all the satellite images for region  \textit{tdr0t}
 over the ten-year period. In more detail:
\begin{itemize}
\item Between $2016$ and $2019$, there is modest growth, with AUI increasing from $2.5$ to $3.5$.
\item Between $2019$ and $2022$, the growth tapers off compared to the previous period, which is reflected in the AUI rising more slowly from $3.5$ to $3.9$.
\item Between $2022$ and $2025$, there is no visible change in urban development, resulting in a flat AUI that climbs very slowly,  from $3.9$ to $4.1$.
\end{itemize}

These examples demonstrate the potential of VLMs for long-term monitoring of urban growth and highlight the limitations of pixel-based indices in doing the same.

\section{Future Work}

The AUI pipeline can be further enhanced by incorporating additional data sources. For example, integrating places data from open sources such as Overture \cite{overturemaps2025} -- which is routinely updated -- can help refine AUI. Furthermore, the AUI time series can be used to forecast future urban development, making it a valuable asset for real estate development, urban planning, and retail strategy.

Finally, the AUI model itself can be further improved through supervised learning. We can use the current VLM-based model to create a large-scale dataset of satellite images with corresponding AUI labels. Once sufficient data is collected, a supervised learning model can be trained on this dataset to compute the AUI  for an image. This process would further simplify AUI computation, offering a more scalable and cost-effective solution.

\section*{Acknowledgement}
The authors thank Aswinpratap Narayanasamy, Sachin Jose Varghese, Shobhit Shukla, Srikrishna Nayak, Sreejan Choudhary, Sumanth N, and Vishaal Rao, their colleagues from Propheus, for their valuable feedback.

\bibliographystyle{plain}
\bibliography{references}  

\clearpage
\appendix
\section{Images Used for Analyzing Region tdr70 (Kempegowda International Airport) – Example 1}\label{s:A}

This appendix shows the time series images used for analyzing region \textbf{tdr70} (Kempegowda International Airport) in Example~1. 
Images span from \textbf{January 2016} to \textbf{January 2025} at six-month intervals.

\begin{figure}[h]
    \centering
    \begin{subfigure}[b]{0.4\linewidth}
        \centering
        \includegraphics[width=\linewidth]{airport-image/sentinel_2016-01-01.jpg}
        \caption{2016-01}
    \end{subfigure}
    \hfill
    \begin{subfigure}[b]{0.4\linewidth}
        \centering
        \includegraphics[width=\linewidth]{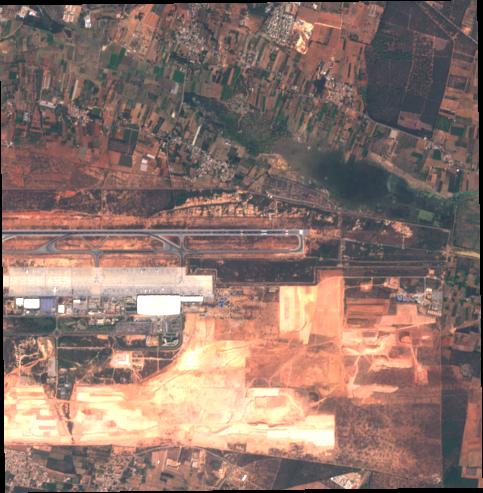}
        \caption{2016-07}
    \end{subfigure}
    
    \vspace{6pt}
    
    \begin{subfigure}[b]{0.4\linewidth}
        \centering
        \includegraphics[width=\linewidth]{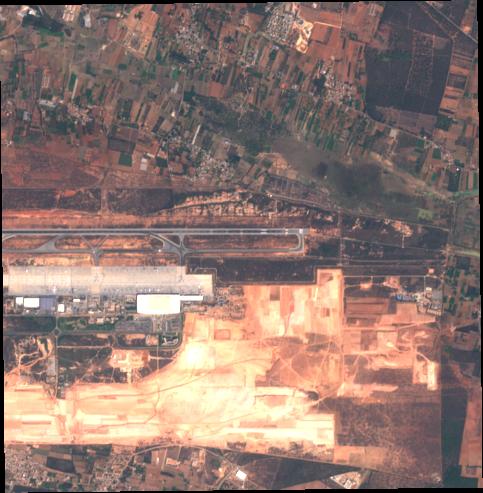}
        \caption{2017-01}
    \end{subfigure}
    \hfill
    \begin{subfigure}[b]{0.4\linewidth}
        \centering
        \includegraphics[width=\linewidth]{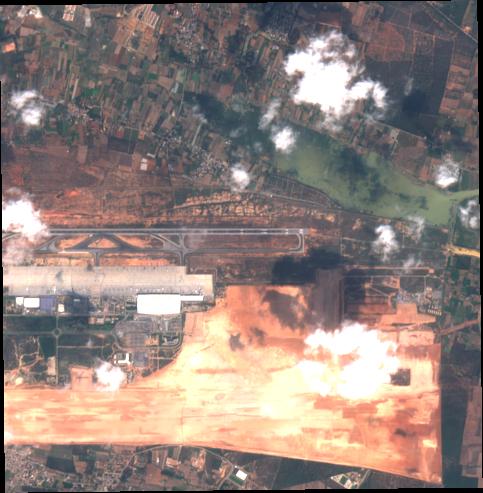}
        \caption{2017-07}
    \end{subfigure}
    
    \caption{Example~1 images for region tdr70 (Kempegowda International Airport), part 1 (Jan 2016 – Jul 2017).}
    \label{fig:exp1_tdr70_subfig_part1}
\end{figure}

\begin{figure}[H]
    \centering
    \begin{subfigure}[b]{0.4\linewidth}
        \centering
        \includegraphics[width=\linewidth]{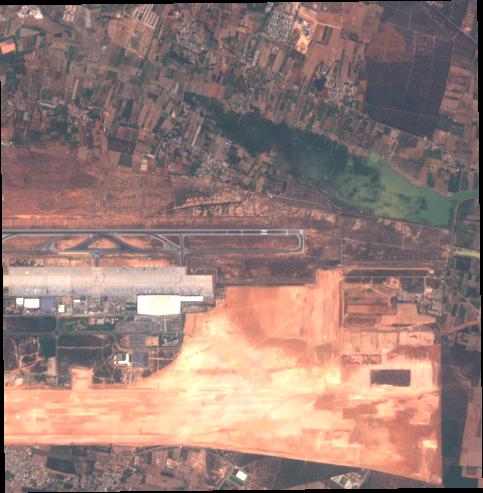}
        \caption{2018-01}
    \end{subfigure}
    \hfill
    \begin{subfigure}[b]{0.4\linewidth}
        \centering
        \includegraphics[width=\linewidth]{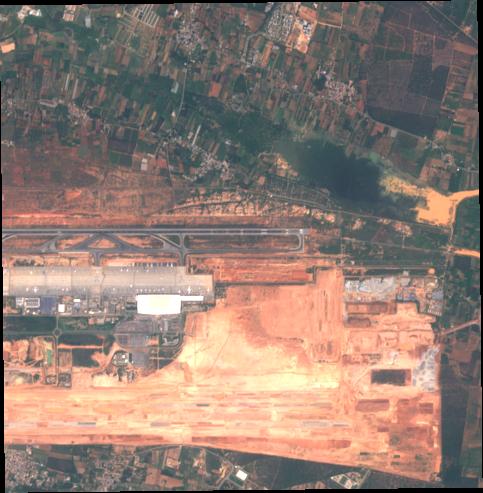}
        \caption{2018-07}
    \end{subfigure}
    
    \vspace{6pt}
    
    \begin{subfigure}[b]{0.4\linewidth}
        \centering
        \includegraphics[width=\linewidth]{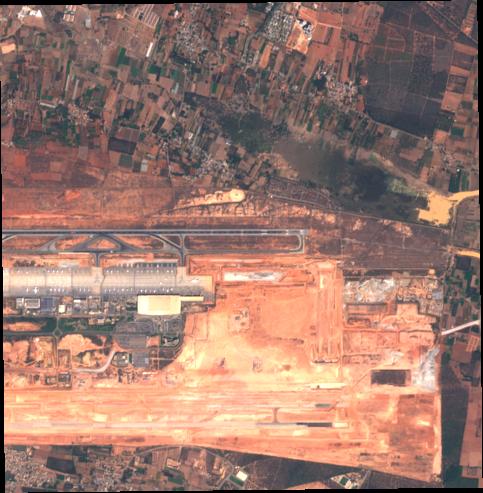}
        \caption{2019-01}
    \end{subfigure}
    \hfill
    \begin{subfigure}[b]{0.4\linewidth}
        \centering
        \includegraphics[width=\linewidth]{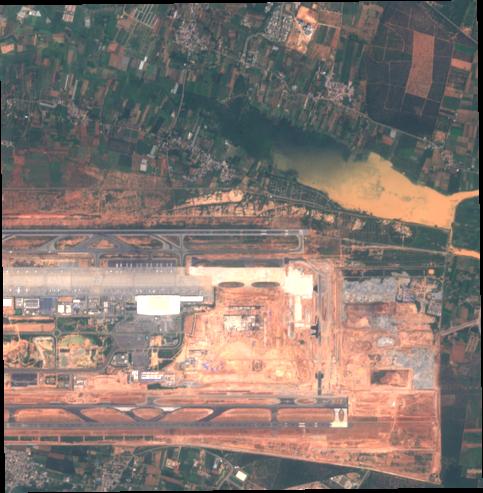}
        \caption{2019-07}
    \end{subfigure}
    
    \vspace{6pt}
    
    \begin{subfigure}[b]{0.4\linewidth}
        \centering
        \includegraphics[width=\linewidth]{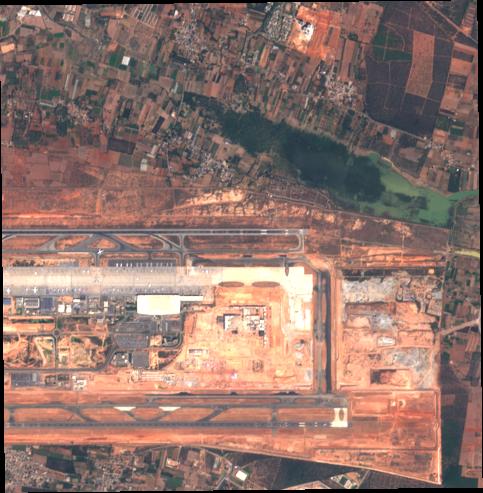}
        \caption{2020-01}
    \end{subfigure}
    \hfill
    \begin{subfigure}[b]{0.4\linewidth}
        \centering
        \includegraphics[width=\linewidth]{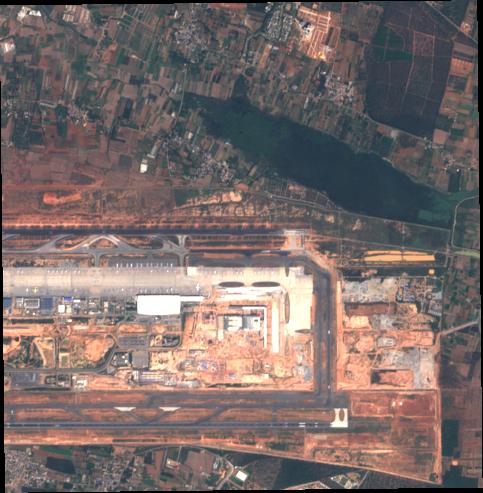}
        \caption{2020-07}
    \end{subfigure}
    
    \caption{Example~1 images for region tdr70, part 2 (Jan 2018 – Jul 2020).}
    \label{fig:exp1_tdr70_subfig_part2}
\end{figure}

\begin{figure}[H]
    \centering
    \begin{subfigure}[b]{0.4\linewidth}
        \centering
        \includegraphics[width=\linewidth]{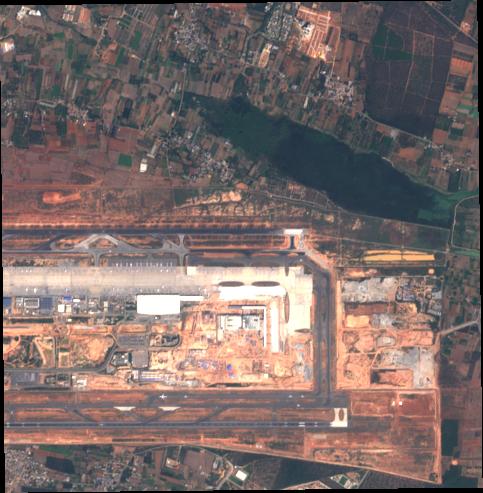}
        \caption{2021-01}
    \end{subfigure}
    \hfill
    \begin{subfigure}[b]{0.4\linewidth}
        \centering
        \includegraphics[width=\linewidth]{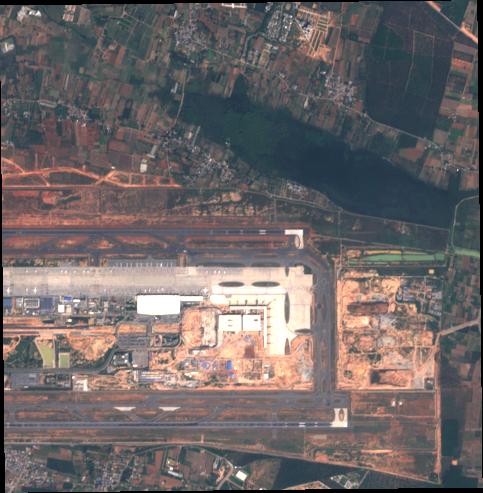}
        \caption{2021-07}
    \end{subfigure}
    
    \vspace{6pt}
    
    \begin{subfigure}[b]{0.4\linewidth}
        \centering
        \includegraphics[width=\linewidth]{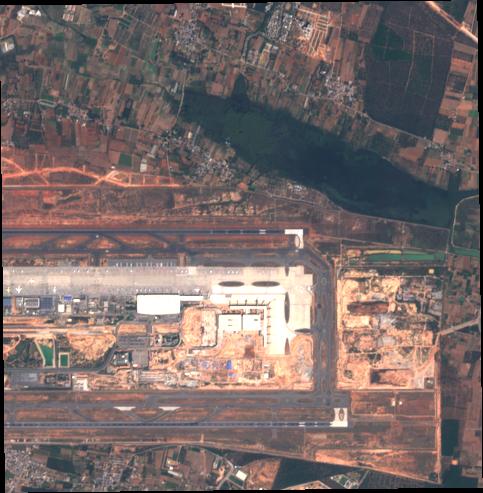}
        \caption{2022-01}
    \end{subfigure}
    \hfill
    \begin{subfigure}[b]{0.4\linewidth}
        \centering
        \includegraphics[width=\linewidth]{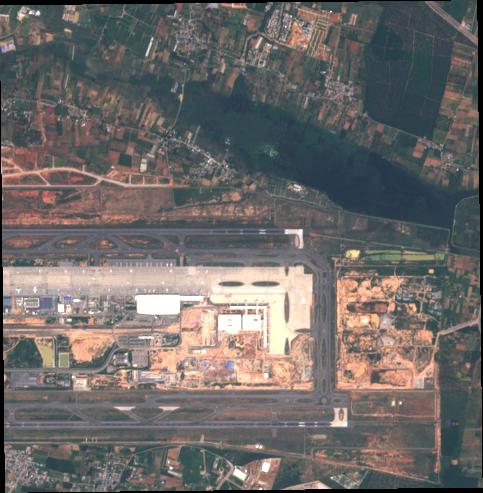}
        \caption{2022-07}
    \end{subfigure}
    
    \vspace{6pt}
    
    \begin{subfigure}[b]{0.4\linewidth}
        \centering
        \includegraphics[width=\linewidth]{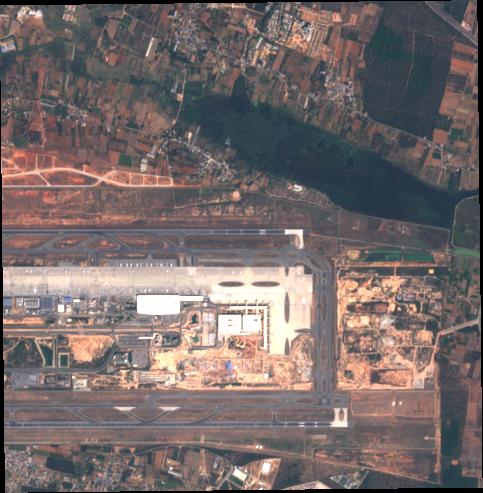}
        \caption{2023-01}
    \end{subfigure}
    \hfill
    \begin{subfigure}[b]{0.4\linewidth}
        \centering
        \includegraphics[width=\linewidth]{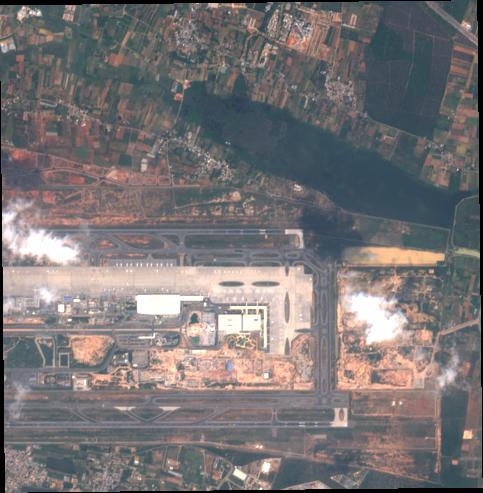}
        \caption{2023-07}
    \end{subfigure}
    
    \caption{Example~1 images for region tdr70, part 3 (Jan 2021 – Jul 2023).}
    \label{fig:exp1_tdr70_subfig_part3}
\end{figure}

\begin{figure}[H]
    \centering
    \begin{subfigure}[b]{0.4\linewidth}
        \centering
        \includegraphics[width=\linewidth]{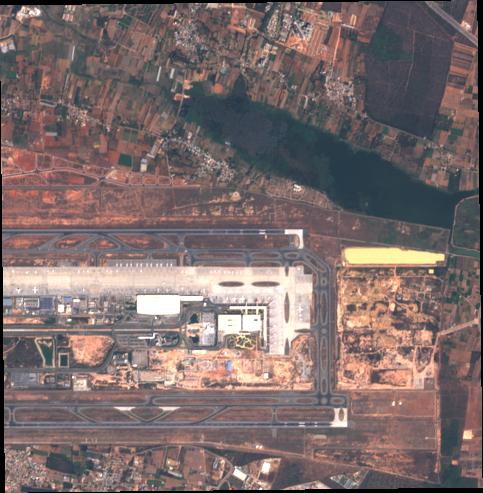}
        \caption{2024-01}
    \end{subfigure}
    \hfill
    \begin{subfigure}[b]{0.4\linewidth}
        \centering
        \includegraphics[width=\linewidth]{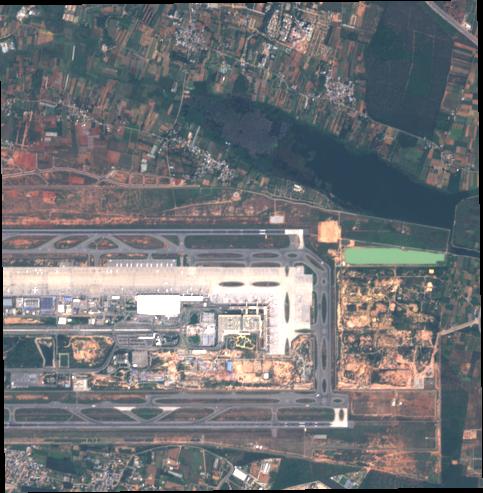}
        \caption{2024-07}
    \end{subfigure}
    
    \vspace{6pt}
    
    \begin{subfigure}[b]{0.4\linewidth}
        \centering
        \includegraphics[width=\linewidth]{airport-image/sentinel_2025-01-01.jpg}
        \caption{2025-01}
    \end{subfigure}
    
    \caption{Example~1 images for region tdr70, part 4 (Jan 2024 – Jan 2025).}
    \label{fig:exp1_tdr70_subfig_part4}
\end{figure}

\clearpage
\section{Images Used for Analyzing Region tdrot (Bannerghatta National Park,) – Example 2}\label{s:B}

This appendix shows the time series images used for analyzing region \textbf{tdrot} (Bannerghatta National Park,) in Example~2. 
Images span from \textbf{January 2016} to \textbf{January 2025} at six-month intervals.

\begin{figure}[h]
    \centering
    \begin{subfigure}[b]{0.4\linewidth}
        \centering
        \includegraphics[width=\linewidth]{bannerghatta-image/sentinel_2016-01-01.jpg}
        \caption{2016-01}
    \end{subfigure}
    \hfill
    \begin{subfigure}[b]{0.4\linewidth}
        \centering
        \includegraphics[width=\linewidth]{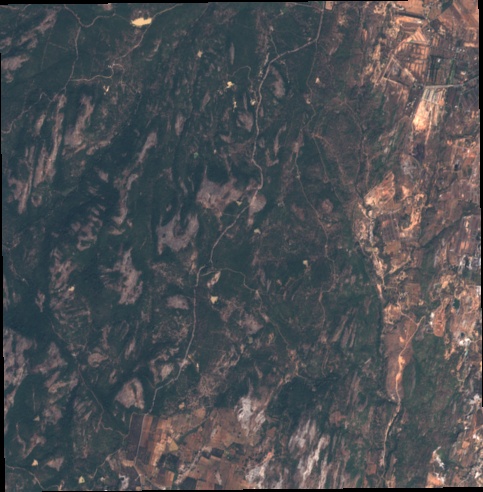}
        \caption{2016-07}
    \end{subfigure}
    
    \vspace{6pt}
    
    \begin{subfigure}[b]{0.4\linewidth}
        \centering
        \includegraphics[width=\linewidth]{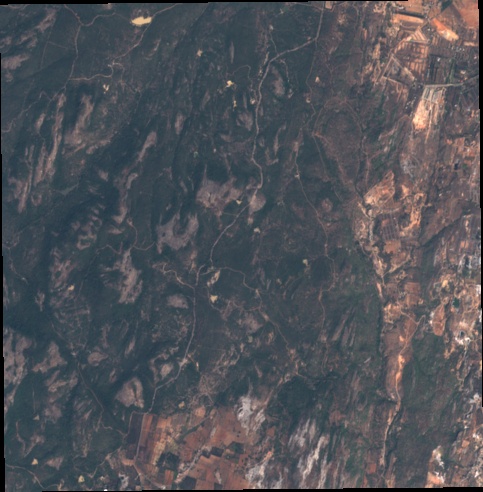}
        \caption{2017-01}
    \end{subfigure}
    \hfill
    \begin{subfigure}[b]{0.4\linewidth}
        \centering
        \includegraphics[width=\linewidth]{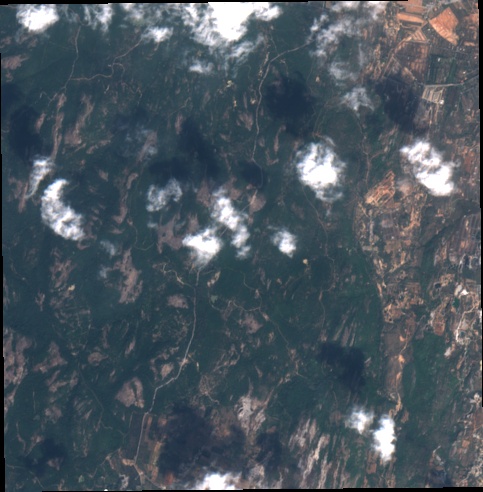}
        \caption{2017-07}
    \end{subfigure}
    
    \caption{Example~2 images for region tdrot (Bannerghatta National Park,), part 1 (Jan 2016 – Jul 2017).}
    \label{fig:exp1_tdrot_subfig_part1}
\end{figure}

\begin{figure}[H]
    \centering
    \begin{subfigure}[b]{0.4\linewidth}
        \centering
        \includegraphics[width=\linewidth]{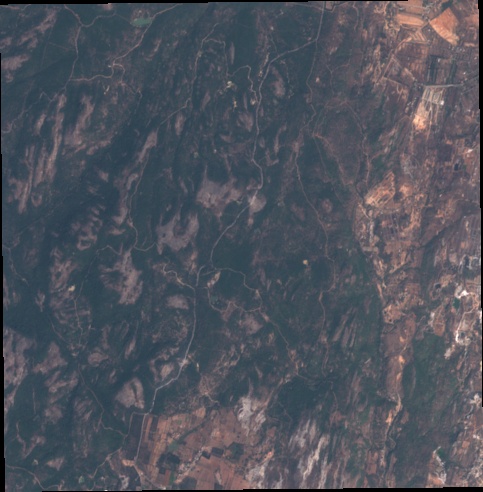}
        \caption{2018-01}
    \end{subfigure}
    \hfill
    \begin{subfigure}[b]{0.4\linewidth}
        \centering
        \includegraphics[width=\linewidth]{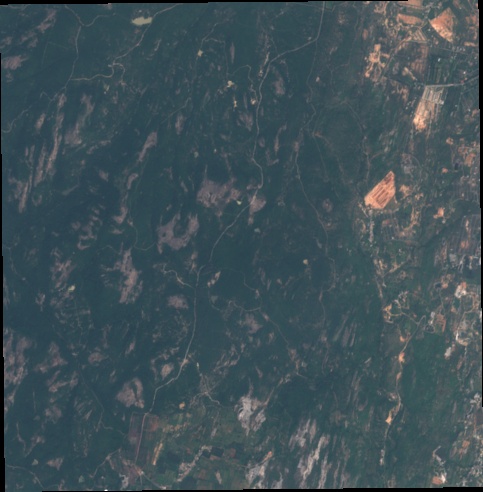}
        \caption{2018-07}
    \end{subfigure}
    
    \vspace{6pt}
    
    \begin{subfigure}[b]{0.4\linewidth}
        \centering
        \includegraphics[width=\linewidth]{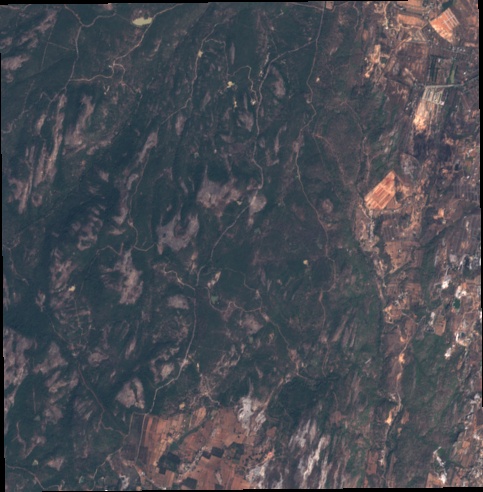}
        \caption{2019-01}
    \end{subfigure}
    \hfill
    \begin{subfigure}[b]{0.4\linewidth}
        \centering
        \includegraphics[width=\linewidth]{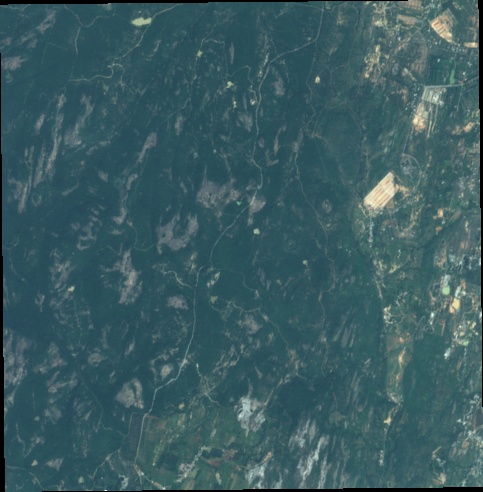}
        \caption{2019-07}
    \end{subfigure}
    
    \vspace{6pt}
    
    \begin{subfigure}[b]{0.4\linewidth}
        \centering
        \includegraphics[width=\linewidth]{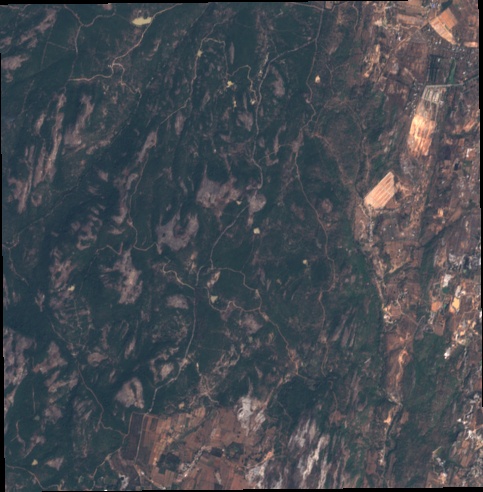}
        \caption{2020-01}
    \end{subfigure}
    \hfill
    \begin{subfigure}[b]{0.4\linewidth}
        \centering
        \includegraphics[width=\linewidth]{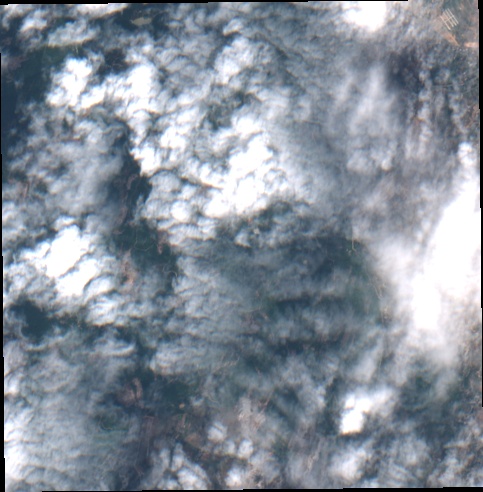}
        \caption{2020-07}
    \end{subfigure}
    
    \caption{Example~2 images for region tdrot, part 2 (Jan 2018 – Jul 2020).}
    \label{fig:exp1_tdrot_subfig_part2}
\end{figure}

\begin{figure}[H]
    \centering
    \begin{subfigure}[b]{0.4\linewidth}
        \centering
        \includegraphics[width=\linewidth]{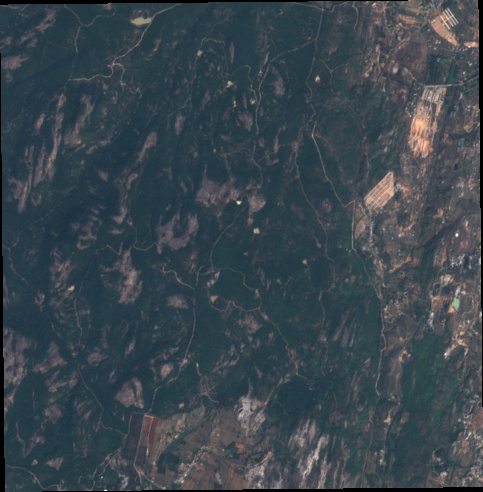}
        \caption{2021-01}
    \end{subfigure}
    \hfill
    \begin{subfigure}[b]{0.4\linewidth}
        \centering
        \includegraphics[width=\linewidth]{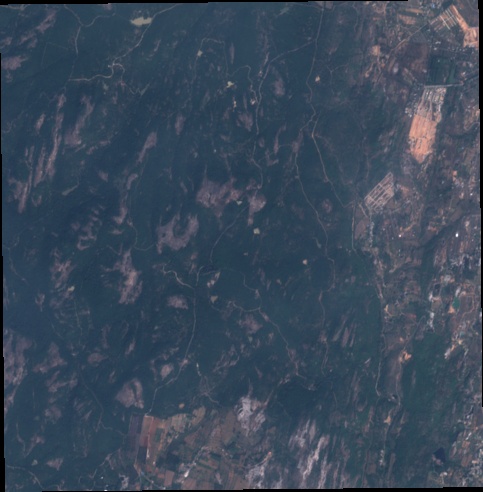}
        \caption{2021-07}
    \end{subfigure}
    
    \vspace{6pt}
    
    \begin{subfigure}[b]{0.4\linewidth}
        \centering
        \includegraphics[width=\linewidth]{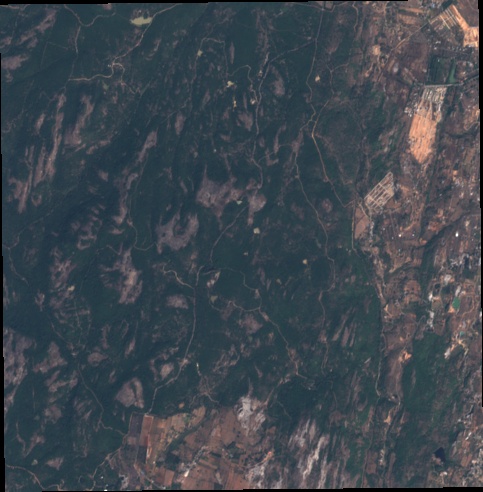}
        \caption{2022-01}
    \end{subfigure}
    \hfill
    \begin{subfigure}[b]{0.4\linewidth}
        \centering
        \includegraphics[width=\linewidth]{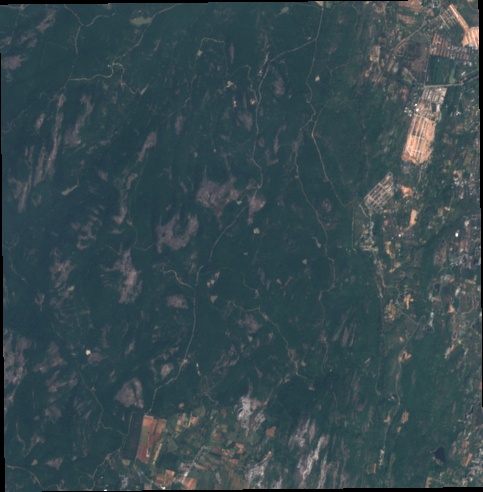}
        \caption{2022-07}
    \end{subfigure}
    
    \vspace{6pt}
    
    \begin{subfigure}[b]{0.4\linewidth}
        \centering
        \includegraphics[width=\linewidth]{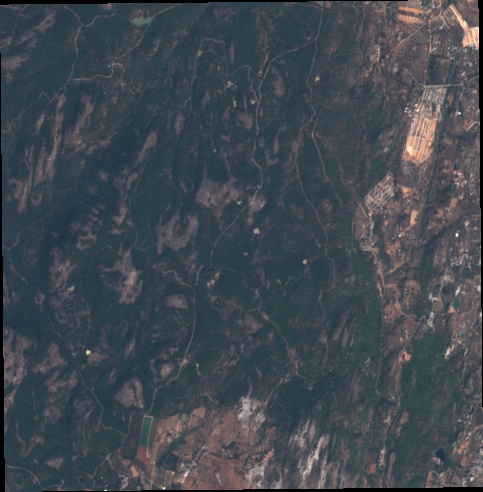}
        \caption{2023-01}
    \end{subfigure}
    \hfill
    \begin{subfigure}[b]{0.4\linewidth}
        \centering
        \includegraphics[width=\linewidth]{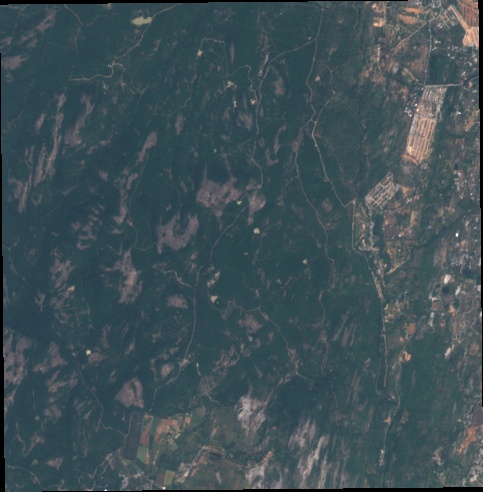}
        \caption{2023-07}
    \end{subfigure}
    
    \caption{Example~2 images for region tdrot, part 3 (Jan 2021 – Jul 2023).}
    \label{fig:exp1_tdrot_subfig_part3}
\end{figure}

\begin{figure}[H]
    \centering
    \begin{subfigure}[b]{0.4\linewidth}
        \centering
        \includegraphics[width=\linewidth]{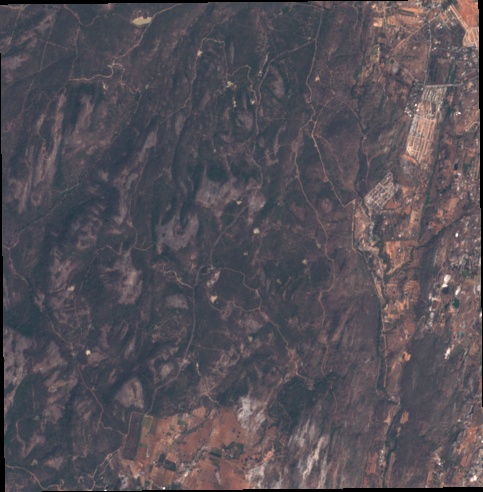}
        \caption{2024-01}
    \end{subfigure}
    \hfill
    \begin{subfigure}[b]{0.4\linewidth}
        \centering
        \includegraphics[width=\linewidth]{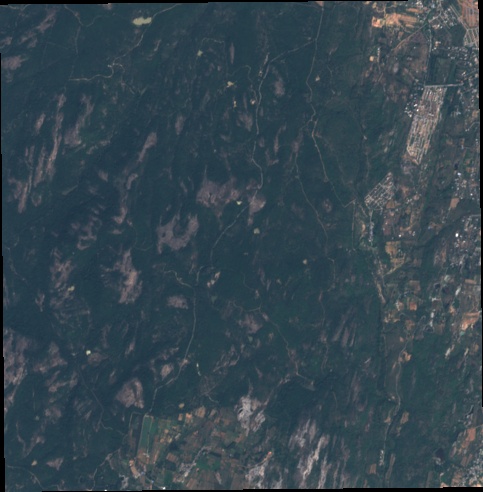}
        \caption{2024-07}
    \end{subfigure}
    
    \vspace{6pt}
    
    \begin{subfigure}[b]{0.4\linewidth}
        \centering
        \includegraphics[width=\linewidth]{bannerghatta-image/sentinel_2025-01-01.jpg}
        \caption{2025-01}
    \end{subfigure}
    
    \caption{Example~2 images for region tdrot, part 4 (Jan 2024 – Jan 2025).}
    \label{fig:exp1_tdrot_subfig_part4}
\end{figure}






\end{document}